\pdfoutput=1

\documentclass[11pt]{article}

\usepackage{acl}

\usepackage{times}
\usepackage{latexsym}

\usepackage[T1]{fontenc}

\usepackage[utf8]{inputenc}
\usepackage{CJKutf8}

\usepackage{xcolor}
\usepackage{microtype}
\usepackage{inconsolata}
\usepackage{graphicx}
\usepackage{bm}
\usepackage{enumitem}
\usepackage{multirow}
\usepackage{amsfonts}
\usepackage{amssymb}
\usepackage{amsmath}
\usepackage{verbatim}
\usepackage{arydshln}
\usepackage{wasysym}
\usepackage{tcolorbox}
\usepackage{algorithm}
\usepackage{algpseudocode}
\usepackage{cleveref}
\usepackage{booktabs}
\usepackage{longtable}
\usepackage{soul}
\sethlcolor{white} 

\definecolor{confHigh}{HTML}{B7E1CD}   
\definecolor{confMed}{HTML}{FFF2B2}    
\definecolor{confLow}{HTML}{F9C4A5}    

\newcommand{\confspan}[3]{%
  \begingroup
    \setlength{\fboxsep}{0.6pt}%
    \colorbox{#1!40}{\strut #2} (#3)%
  \endgroup
}

\newcommand{\zh}[1]{{\begin{CJK*}{UTF8}{gbsn}#1\end{CJK*}}}
\newcommand{\datasetname}{\textsc{FiNERweb}}

\title{\datasetname{}: Datasets and Artifacts for \\ Scalable Multilingual Named Entity Recognition}

\author{
    Jonas Golde \hspace{1em} Patrick Haller  \hspace{1em} Alan Akbik \\
    Humboldt Universität zu Berlin \\
    \texttt{jonas.max.golde.1@hu-berlin.de}
}

\begin{document}
\maketitle

\begin{abstract}
Recent multilingual named entity recognition (NER) work has shown that large language models (LLMs) can provide effective synthetic supervision, yet such datasets have mostly appeared as by-products of broader experiments rather than as systematic, reusable resources. We introduce \datasetname{}, a dataset-creation pipeline that scales the teacher–student paradigm to 91 languages and 25 scripts. Building on FineWeb-Edu, our approach trains regression models to identify NER-relevant passages and annotates them with multilingual LLMs, resulting in about 225k passages with 235k distinct entity labels. Our experiments show that the regression model achieves more than 84 F1, and that models trained on \datasetname{} obtain comparable or improved performance in zero shot transfer settings on English, Thai, and Swahili, despite being trained on 19x less data than strong baselines. In addition, we assess annotation quality using LLM-as-a-judge and observe consistently high scores for both faithfulness (3.99\,/\,5) and completeness (4.05\,/\,5), indicating reliable and informative annotations. Further, we release the dataset with both English labels and translated label sets in the respective target languages because we observe that the performance of current state-of-the-art models drops by 0.02--0.09 F1 when evaluated using target language labels instead of English ones. We release \datasetname{} together with all accompanying artifacts to the research community in order to facilitate more effective student-teacher training for multilingual named entity recognition.
\end{abstract}

\section{Introduction} \label{sec:introduction}
Named entity recognition (NER) is the task of identifying tokens that belong to a predefined set of classes such as ``person'' or ``location'' \citep{lample-etal-2016-neural,akbik-etal-2018-contextual}. In recent years, advances in large language models (LLMs) have enabled zero-shot NER across domains and label types \citep{xie-etal-2024-self,wang-etal-2025-gpt}. However, the paradigm of prompting is not an ideal fit for the task of information extraction: it requires issuing separate prompts for each label type, or the generated annotations need postprocessing to align with the original text \citep{santoso-etal-2024-pushing}. As a consequence, recent work has leveraged LLMs to generate synthetic training data for smaller, more efficient models such as UniNER \citep{zhou2024universalner}, LitSet \citep{golde-etal-2024-large}, and GLiNER \citep{zaratiana-etal-2024-gliner}. These models effectively distill knowledge from LLMs and, despite being orders of magnitude smaller, can outperform their larger teacher models, establishing the state-of-the-art in NER.

However, while autoregressive LLMs such as Qwen3 \citep{yang2025qwen3technicalreport}, Gemma3 \citep{gemmateam2025gemma3technicalreport}, and GPT-4 \citep{openai2024gpt4technicalreport} support more than 100 languages, current universal NER models focus mostly on the English language, and no multilingual dataset currently exists to train those models. Existing human-labeled multilingual NER resources also fail to close this gap because they typically provide only one of the two necessary dimensions: (\textit{i}) broad language coverage or (\textit{ii}) a rich label set. For instance, datasets such as PAN-X \citep{pan-etal-2017-cross} cover over 150 languages but restrict the label set to coarse types such as ``person'', ``location'', and ``organization'' whereas datasets such as DynamicNER \citep{luo2025dynamicnerdynamicmultilingualfinegrained} support over 100 label types but only support 12 comparatively high-resource languages.

\begin{figure*}
    \centering
    \includegraphics[width=\linewidth]{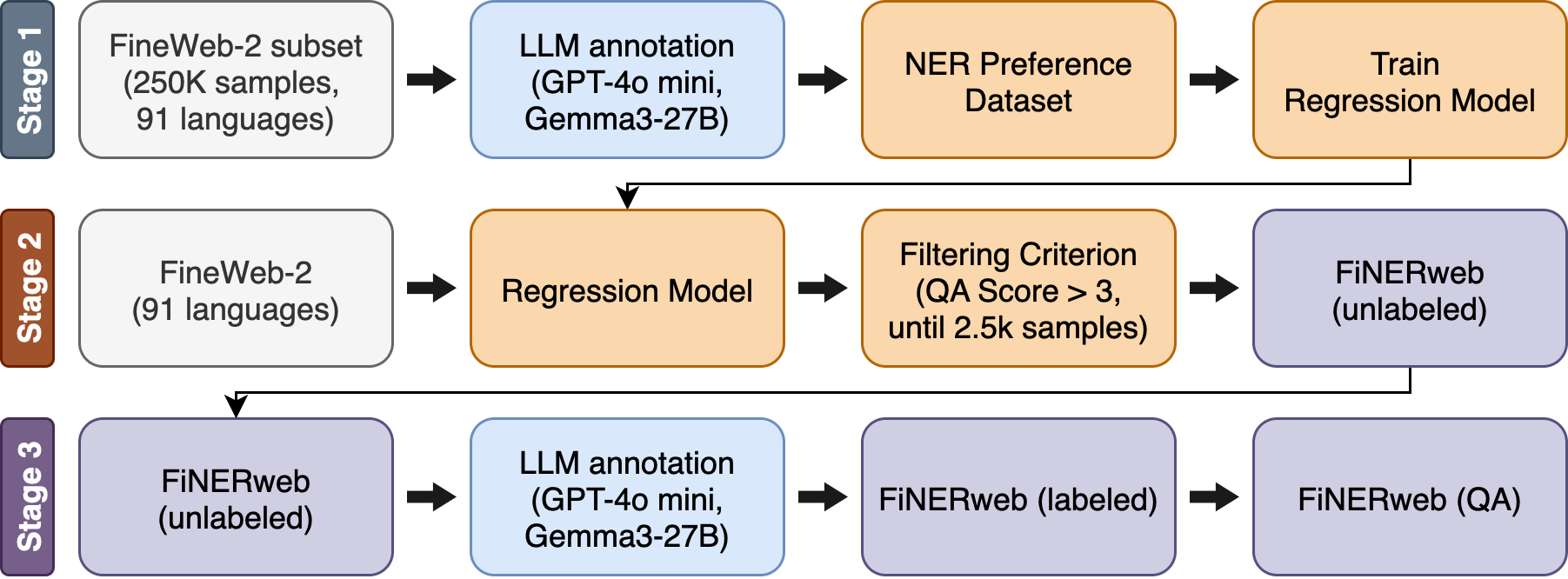}
    \caption{Three-stage pipeline for constructing \datasetname{}: We (1) use a multilingual LLM to generate preference data for identifying high-quality NER examples, (2) train a regression model to filter the FineWeb2 corpus, and (3) annotate the filtered passages with a multilingual LLM.}
    \label{fig:stages_explanation}
\end{figure*}

In this paper, we introduce a scalable dataset creation pipeline for multilingual named entity recognition (NER) using large language models (LLMs). The pipeline builds on the FineWeb-Edu methodology \citep{penedo2024finewebdatasetsdecantingweb} and combines automated passage selection and LLM-based annotation to create the dataset. As the outcome of this pipeline, we release \datasetname{}, an LLM-annotated dataset covering 91 languages and 25 scripts. 

The pipeline consists of several steps: First, we apply an LLM to rate the usefulness of 1k passages per language for the NER task. These ratings are used to train a multilingual regression model based on XLM-RoBERTa \citep{conneau-etal-2020-unsupervised}. We use the resulting regression model to select 2.5k passages per language from the FineWeb-2 corpus \citep{penedo2025fineweb2pipelinescale}. We annotated the selected passages using GPT-4o mini and Gemma3-27B and merge the NER annotations into a single annotation set. At last, we translate all English annotations into the corresponding target languages using the Google Translate API \citep{google_translation_api_v3}.

In our experiments, we demonstrate the usefulness of the resulting dataset by (i) showing that we can train models on selected splits of our dataset and achieve comparable or improved performance compared to current state-of-the-art models and (ii) evaluate the quality of the resulting dataset using LLM-as-a-judge with Qwen3-235B-A22B \citep{yang2025qwen3technicalreport}. In addition, we analyze the effect of translating the label set into the respective target languages and investigate the long-tail distribution of entity types of the resulting dataset. We summarize our main contributions as follows:

\begin{enumerate}
    \item We present a scalable pipeline for constructing multilingual NER datasets that combines automated passage selection, LLM-based annotation, and label translation.
    \item As the outcome of this pipeline, we release \datasetname{}, an LLM-annotated dataset covering 91 languages and 25 scripts, together with the regression models and their training data.
    \item We conduct extensive experiments to show to usefulness of our dataset including downstream training and LLM-as-a-judge evaluation.
    \item We release all artifacts to the research community.\footnote{\url{https://github.com/whoisjones/fiNERweb}}
\end{enumerate}

\section{FiNERweb} \label{sec:dataset_creation}

In this section, we present the three-stage process used to derive \datasetname{}.~\Cref{fig:stages_explanation} illustrates the steps involved. First, we obtain preference training data to train a regression model, which we subsequently apply to filter high-quality passages suitable for NER annotation, e.g., excluding advertisements and other irrelevant content. Finally, we annotate the filtered dataset using two LLMs and merge their outputs to maximize coverage and correctness of entity types.

\begin{table*}[h]
\centering
\setlength{\tabcolsep}{6pt}
\renewcommand{\arraystretch}{1.1}
\begin{tabular}{lllccc}
\toprule
\textsc{Task} & \textsc{Annotation LLM} & \textsc{Transformer} & \textsc{Precision} & \textsc{Recall} & \textsc{F1} \\
\midrule
\multirow{4}{*}{binary}
  & \multirow{2}{*}{GPT-4o mini} & XLM-RoBERTa & \underline{0.841} & \textbf{0.842} & \textbf{0.841} \\
  &                         & mDeBERTa    & 0.821 & \underline{0.827} & \underline{0.823} \\
\cmidrule(lr){2-6}
  & \multirow{2}{*}{Gemma3 27B}  & XLM-RoBERTa & 0.788 & 0.761 & 0.773 \\
  &                         & mDeBERTa    & \textbf{0.849} & 0.536 & 0.504 \\
\midrule
\multirow{4}{*}{multi-class}
  & \multirow{2}{*}{GPT-4o mini} & XLM-RoBERTa & \textbf{0.667} & \textbf{0.666} & \textbf{0.656} \\
  &                         & mDeBERTa    & 0.582 & 0.559 & 0.560 \\
\cmidrule(lr){2-6}
  & \multirow{2}{*}{Gemma3 27B}  & XLM-RoBERTa & \underline{0.637} & \underline{0.570} & \underline{0.569} \\
  &                         & mDeBERTa    & 0.595 & 0.522 & 0.515 \\
\bottomrule
\end{tabular}
\caption{Evaluation results (macro-averaged precision, recall and F1 across languages) of our regression models trained on data annotated by different LLMs (4o-mini and Gemma3-27B) and using different transformer architectures. We highlight the best results per setting in bold and the second-best with underlining.}
\label{tab:preference_classifier_results}
\end{table*}

\begin{table*}[t]
\centering
\small
\begin{tabular}{p{0.46\textwidth} p{0.46\textwidth}}
\toprule
\textsc{High-quality example (score > 3.5)} & \textsc{Low-quality example (score < 0.5)} \\
\midrule
Kraft Foods has taken the Cadbury chocolate brand in a new direction, by combining it with cheese for the first time. The company is bringing together two of its brands and launching Philadelphia with Cadbury, a chilled chocolate spread made from Philadelphia Light and Cadbury chocolate. [...] & Viewing Single Post From: Spoilers for the Week of February 11th. Don't care about Chloe/Taniel/Jen-Jen. Don't care about Sami, really, but hoping that we get some good “SAMANTHA GENE!!” Marlena Death-Stares out of it. STEFANO!! STEFANO, STEFANO, STEFANO!!! [...] \\
\bottomrule
\end{tabular}
\caption{English passages selected by our regression model. The high quality passage includes a richer set of entity types, many inferable from context, whereas the low quality passage contains no named entities.}
\label{tab:good-bad-examples}
\end{table*}

\subsection{Dataset Source}
We select FineWeb-2 as our source dataset which covers more than 1,000 languages. However, we cannot simply include all languages, since our work is constrained by the multilingual capabilities of available language models. Therefore, we restrict ourselves to the languages used to train XLM-RoBERTa. This choice ensures compatibility, as the resulting dataset can be directly applied to train or fine-tune XLM-RoBERTa or similar models such as mDeBERTa~\citep{he2021debertadecodingenhancedbertdisentangled}. Moreover, we assume that the languages used for XLM-RoBERTa pretraining are also likely to be supported in more recent multilingual LLMs such as Qwen3 or Gemma3.

\begin{figure*}
    \includegraphics[width=\textwidth]{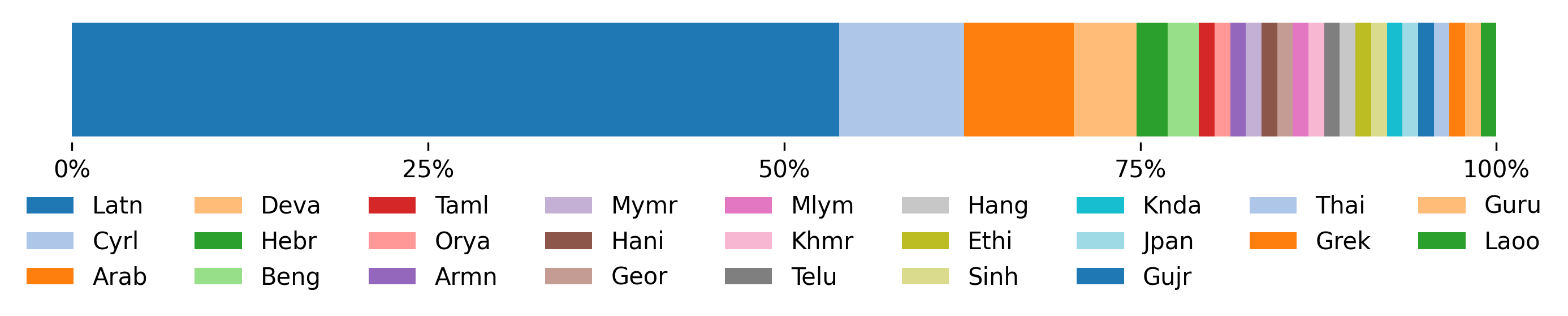}
    \caption{Distribution of scripts in \datasetname{}: Around 50\% of the languages use the Latin script, while the dataset covers a long tail of 25 different scripts in total.}
    \label{fig:script_distribution}
\end{figure*}

Further, we use each language in its native type script (for example, Arabic in Arabic script rather than Latin transliteration), and filtered out languages not shared between XLM-R and FineWeb-2. This leaves us with 91 languages in 25 type scripts. As shown in Figure~\ref{fig:script_distribution}, the distribution is dominated by Latin script, followed by smaller shares for scripts such as Cyrillic and Arabic, with a long tail of lower-resource scripts such as Thai and Khmer.

\subsection{Stage 1: Selecting High-Quality Passages}
In the first step, we aim to filter out noisy and irrelevant texts, such as advertisements or passages that do not contain a diverse range of entity types. This step is necessary because FineWeb-2 is sourced from 96 CommonCrawl snapshots and thus possibly contains many low-quality data points for NER. To do so, we train a regression model that can score the potential quality of a passage for NER annotation.

To construct the training data for the regression model, we randomly sample 1k examples per language. Each example is obtained by splitting a data point into chunks of 256 tokens using word segmentation models of spaCy \citep{honnibal2020spacy}, Janome \citep{janome} and Stanza \citep{qi2020stanzapythonnaturallanguage} from which we sample exactly one chunk to increase diversity. We then use GPT-4o mini and Gemma3-27B to rate the usefulness of each passage for NER on a scale from 1~to~4, following the prompt shown in~\Cref{fig:preference_prompt}. The scale of scores is defined as:

\begin{itemize}
    \item \textbf{1 point:} None, few or ambiguous entities with little context for reliable identification.
    \item \textbf{2 points:} Clean and coherent text with identifiable entities and minimal noise.
    \item \textbf{3 points:} Diverse entities across domains with sufficient contextual clues.
    \item \textbf{4 points:} Rich, well-contextualized, and noise-free text suitable for NER training and evaluation.
\end{itemize}

We show example ratings in \Cref{tab:good-bad-examples} for English and in \Cref{tab:examples-other-languages} for additional languages.

\begin{figure*}
    \centering
    \includegraphics[width=\linewidth]{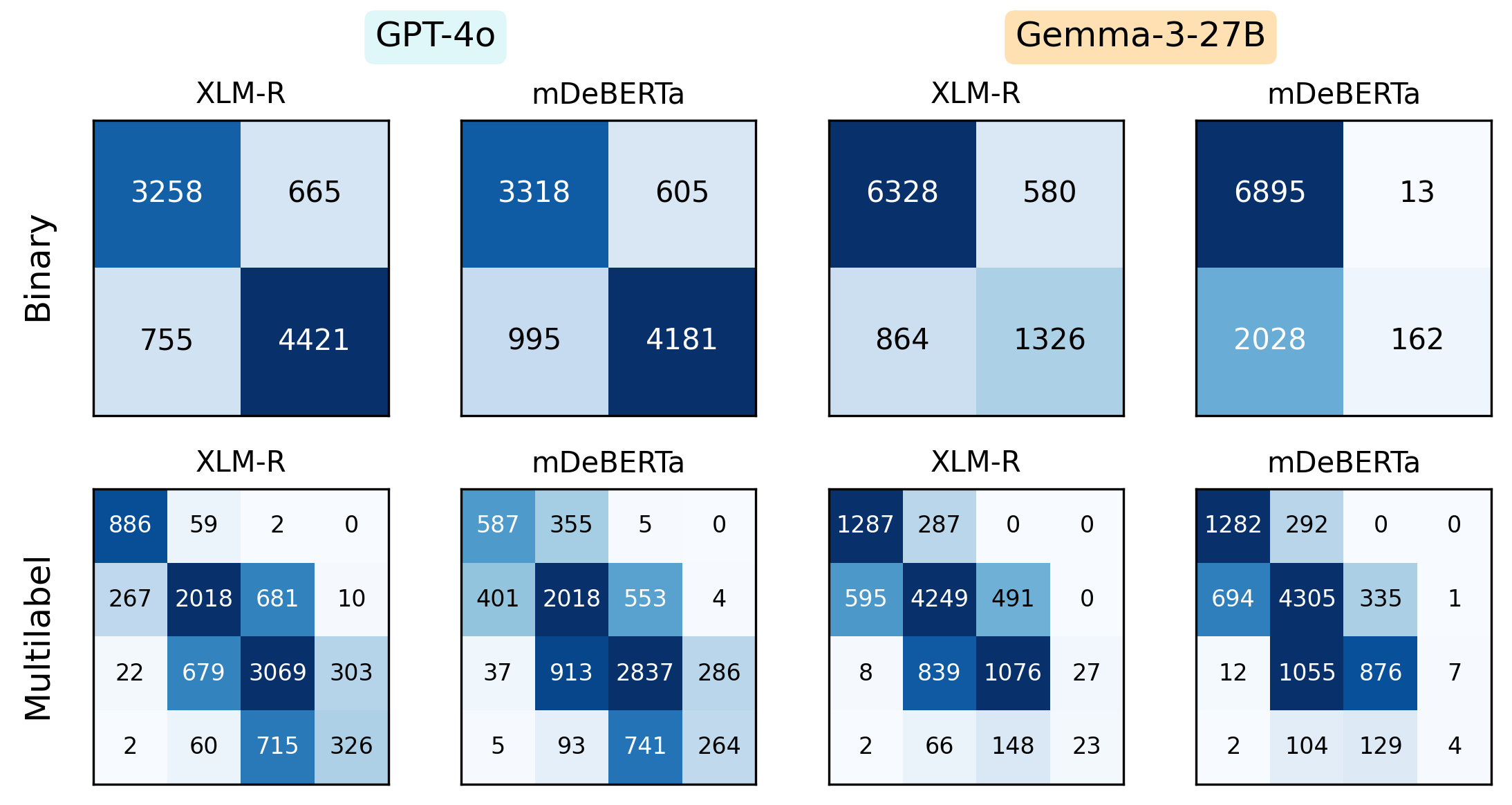}
    \caption{Confusion matrices for regression models reported in~\Cref{tab:preference_classifier_results} showing the distribution of prediction errors. GPT-4o mini annotations yield more balanced predictions along the diagonal, whereas Gemma3 annotations rarely give the highest score. Most errors occur close to the diagonal, indicating models avoid severe misclassifications.}
    \label{fig:preference_classifier_cm}
\end{figure*}

\noindent\textbf{Regression Model Training.}~Let $\mathcal{D} = \{(x_i, y_i)\}_{i=1}^N$ denote the annotated preference dataset, where $x_i$ is a passage and $y_i \in \{1,2,3,4\}$ is the usefulness score assigned by the LLM. We employ a simple regression model $\theta$ to encode $x_i$ and obtain a predicted score $\hat{y}_i \in \mathbb{R}$, which we optimize by minimizing the mean squared error (MSE) between predictions and labels:  

\[
\mathcal{L}_{\text{reg}} = \frac{1}{N} \sum_{i=1}^N \left( y_i - \hat{y}_i \right)^2.
\]

\noindent\textbf{Experimental Setup.}~We employ two LLMs to annotate the random 1k passages per language, namely GPT-4o mini and Gemma3-27B, and experiment with two different multilingual transformer architectures: XLM-RoBERTa and mDeBERTa. For both models, we use a learning rate of $5e^{-5}$ and a batch size of 64, training for up to 20 epochs with early stopping (patience of 5) in half precision. We allocate 10\% of the data as a validation split, using a stratified sample across languages.

We further train two variants: (i) a binary model, in which the models learn to distinguish between useful and non-useful passages (score $>= 3$), and (ii) a multi-class variant, in which the model is trained on all four categories to enable more fine-grained filtering. For each setting, we train three models using different seeds and report the macro-averaged results over languages across these runs.

\noindent\textbf{Results.}~We present the classification results of regression model in~\Cref{tab:preference_classifier_results} by mapping the continuous predictions $\hat{y}_i$ back to discrete labels by rounding to the nearest integer. We observe that XLM-RoBERTa consistently outperforms mDeBERTa, achieving for instance +2.2 F1 in the binary setting and +8.7 F1 in the multi-class setting when using GPT-4o mini annotations. A similar trend holds for annotations provided by Gemma3-27B. Moreover, models trained on GPT-4o mini annotations consistently yield higher performance compared to those trained on Gemma3 annotations, e.g., 84.1 F1 (binary / GPT-4o-mini annotations / XLM-RoBERTa transformer) versus 77.3 F1 (binary / Gemma3 annotations / XLM-RoBERTa transformer).

We further analyze the model errors in~\Cref{fig:preference_classifier_cm}. GPT-4o mini annotations are balanced across the diagonal of the confusion matrix, whereas Gemma3-27b annotations are skewed, rarely assigning the maximum score of 4. Although not perfect, most regression model errors are concentrated near the diagonal, suggesting that the models avoid severe misclassifications. In contrast, Gemma3 annotations classify only a small number of passages as highly useful. Based on these findings, we select the binary XLM-RoBERTa regression model trained on GPT-4o mini annotations for Stage~2 of our process.

\begin{table*}[t]
\centering
\begin{tabular}{lccc|ccc}
\toprule
 & \multirow{2}{*}{\textsc{NuNER}} & \multirow{2}{*}{\textsc{PileNER}} & \textsc{Euro-} & \multicolumn{3}{c}{\textsc{FiNERWeb}} \\
 &  &  & \textsc{GLiNER-X} & 4o-mini & Gemma & Merged \\
\midrule
\# Samples & \textbf{968.4k} & 45.9k & 85.2k & 226.0k & 225.6k & 226.0k \\
Avg. Text Length & 147.8 & 1063.7 & \textbf{1501.9} & 1310.9 & 1308.1 & 1310.9 \\
Avg. Types/Sample & 4.5 & 20.5 & 8.4 & 19.4 & 21.5 & \textbf{25.4} \\
Avg. Unique Types/Sample & 3.3 & 5.2 & 3.7 & 6.6 & 9.5 & \textbf{11.9} \\
Unique Entity Types & 192k & 12.6k & 20.0k & 28.4k & 134k & \textbf{235k} \\
\# Languages & 1 & 1 & 14 & \textbf{91} & \textbf{91} & \textbf{91} \\
\bottomrule
\end{tabular}
\caption{Comparison of \datasetname{} with universal NER datasets NuNER, PileNER, and Euro-GLiNER-X. We highlight the largest quantities in bold.}
\label{tab:dataset_stats}
\end{table*}

\subsection{Stage 2: Filtering FineWeb-2}
With the regression model from Stage~1, we filter FineWeb-2 for high-quality examples in each target language. We apply identical pre-processing and split each example into chunks of 256 tokens. We then compute the usefulness score using our regression model. If the predicted score is $> 0.5$ (which is equivalent to the passage scoring $>= 3$). Once we found a useful chunk, we apply an additional quality-control step using a fine-tuned language identification model\footnote{\url{https://huggingface.co/qanastek/51-languages-classifier}} to detect whether a passage is not in English as we found some websites to contain embedded English advertisement texts that would be considered useful by our model. If this step is passed, we add this chunk to our unlabeled dataset and proceed with the next data point to increase diversity of our dataset.

\begin{table*}[h!]
\centering
\renewcommand{\arraystretch}{1.1}
\begin{tabular}{lcccr}
\hline
\multirow{2}{*}{\textsc{Model}} & \textsc{CoNLL} & \textsc{MaskahaNER} & \textsc{ThaiNER} & \multirow{2}{*}{\textsc{Avg.}}  \\
& \textsc{eng} & \textsc{swa} & \textsc{tha}  \\
\hline
\textit{Fine-tuned on PileNER} \\
GLiNER-multi-v2.1 & 0.601 & 0.642 & \textbf{0.578} & 0.607 \\
\hline
\textit{Fine-tuned on Euro-GLiNER-X} \\
GLiNER-X-base & \textbf{0.671} & \underline{0.764} & 0.431 & \textbf{0.622} \\
\midrule
\textit{Fine-tuned on \datasetname{}} \\
Each target language (\textsc{eng, swa}, or \textsc{tha}) & 0.585 & 0.680 & \underline{0.532} & 0.600 \\
All target languages (\textsc{eng, swa}, and \textsc{tha}) & \underline{0.660} & \textbf{0.770} & 0.420 & \underline{0.615} \\
\hline
\end{tabular}
\caption{Results when fine-tuning Binder on selected splits of \datasetname{}. We highlight best scores in bold and second-best underlined.}
\label{tab:downstream_results}
\end{table*}

\subsection{Stage 3: Dataset Annotation}
The resulting unlabeled dataset contains 2.5k passages per language. We then use GPT-4o mini and Gemma3-27B to annotate these passages. Specifically, we instruct each model to generate a list of tuples, where each tuple consists of an entity mention and its corresponding type. We show the corresponding prompt in~\Cref{fig:entity_annotation_prompt}.

\noindent\textbf{Aligning Text and Annotations.} To align the LLM-generated annotations with the original text, we post-process the model outputs as follows: First, we require that every generated entity mention must be an exact substring of the input passage. Annotations that do not satisfy this constraint are discarded. Second, annotations are processed sequentially: the position of the last matched entity determines the starting point for the next match. If a subsequent entity is predicted to occur earlier in the passage (i.e., before the current counter), it is discarded. If we do not apply this constraint, we would propagate annotations through the entire document, resulting in more annotations than the LLM originally made. For example for Gemma3-27b, we keep 77.4\% of annotations. This ensures consistent left-to-right annotation, as enforced by our model instruction. The matching algorithm used for this step is described in~\Cref{alg:extract_spans}.

\noindent\textbf{Semantic Merging of Entity Types.} We merge the annotations produced by GPT-4o mini and Gemma3-27B as follows. For each example, we compare the annotations generated by both models. When span boundaries overlap by less than 50\%, we retain the longer span and discard the shorter one. If no boundary overlap exists, we keep the annotation from the respective model.

For all remaining cases, where span boundaries overlap by at least 50\%, we compute the semantic similarity between the annotation labels using \texttt{all-MiniLM-L6-v2} \citep{reimers-gurevych-2019-sentence}. If the similarity between the annotations exceeds 0.75, we merge them by concatenating their types, for example, ``person'' and ``human'' become ``person / human''.

Overall, 31.5\% of the annotations produced by Gemma3-27B and GPT-4o mini match exactly and are therefore all retained. In total, we retain 66.3\% of the annotations from GPT-4o mini and 60.0\% from Gemma3-27B, corresponding to 63.02\% of all annotations. We further observe that GPT-4o mini tends to produce longer spans on average, which explains why a slightly larger proportion of its annotations is preserved. As a final step, we translate each label set into the respective target language whenever supported by Google Translate. At last, we use Segment-Any-Text \citep{frohmann2024segmenttextuniversalapproach}, a neural sentence segmentation model, to split the passages into single sentences.

We show an overview of \datasetname{} comparing it to other universal NER datasets in~\Cref{tab:dataset_stats}, namely NuNER \citep{bogdanov-etal-2024-nuner}, PileNER \citep{zhou2024universalner}, and Euro-GLiNER-X\footnote{\url{https://huggingface.co/datasets/knowledgator/gliner-multilingual-synthetic}}. Our analysis shows that \datasetname{} contains, on average, more (distinct) annotations per sentence than all baselines and covers a larger set of distinct entity types.

\section{Ablations}

\subsection{Downstream Performance}

In order to investigate the usefulness of our dataset, we fine-tune the Binder architecture \citep{zhang2023optimizingbiencodernamedentity} on selected language splits of \datasetname{}, using \texttt{mBERT} \citep{devlin-etal-2019-bert} as the underlying transformer model. We reuse the hyperparameters proposed by \citet{zhang2023optimizingbiencodernamedentity}. Specifically, we fine-tune the model either on the English, Swahili, and Thai splits individually or jointly on all three languages. We then perform zero-shot evaluation on human-labeled datasets in the corresponding target languages, namely CoNLL-2003 \citep{tjong-kim-sang-de-meulder-2003-introduction}, the Swahili split of MasakhaNER \citep{adelani-etal-2022-masakhaner}, and ThaiNER \citep{wannaphong_phatthiyaphaibun_2024_10795907}. We use multilingual GLiNER models as baseline models.

\noindent\textbf{Results.} We present the results in \Cref{tab:downstream_results}. The model achieves comparable performance on CoNLL-2003 and ThaiNER, and improved performance on the Swahili split of MasakhaNER, despite being trained on only 7.5k passages in the multilingual setting, or 2.5k passages in the per-language fine-tuning setting. While we observe a substantial increase in downstream performance for English and Swahili when training on the combined multilingual data, performance on Thai decreases. We attribute this observation to Binder not requiring word-segmented inputs. As a result, the ratio of positive to negative examples differs substantially across English, Swahili, and Thai, causing the model to preferentially optimize for the easier languages. Based on these experiments, we conclude that (i) \datasetname{} is useful for downstream training, and (ii) careful design of downstream architectures is necessary to ensure robustness across the diverse scripts covered by \datasetname{}.

\begin{table*}[!htbp]
\centering
\label{tab:annotation_errors}
\renewcommand{\arraystretch}{1.15}
\begin{tabular}{lccp{8cm}}
\toprule
 & \textsc{Count} & \textsc{\%} & \textsc{Top-5 Entity Types} \\
\midrule
All annotations & 66,329 & -- & -- \\
Missing annotations & 4,062 & 6.12\% & person (28.73\%), event (16.25\%), organization (15.14\%), date (14.18\%), location (14.18\%) \\
Wrong annotations & 3,961 & 5.97\% & cultural reference (24.69\%), person (22.80\%), location (15.60\%), scientific concept (15.30\%), organization (15.15\%) \\
\bottomrule
\end{tabular}
\caption{Error analysis based on LLM as a judge evaluation across all languages.}
\end{table*}

\subsection{Annotation Quality}

We assess annotation quality using an LLM-as-a-judge \citep{zheng2023judging} using Qwen-235B \citep{yang2025qwen3technicalreport}. For each language, we randomly sample 25 examples and evaluate them along two dimensions: \emph{faithfulness}, which measures whether the annotations made are actually correct and \emph{completeness}, which measures how many entities occur in the passage that have not been annotated. Each example is rated on a five point scale following the guidelines shown in \Cref{fig:ner_quality_prompt}.

\begin{figure}
    \includegraphics[width=\linewidth]{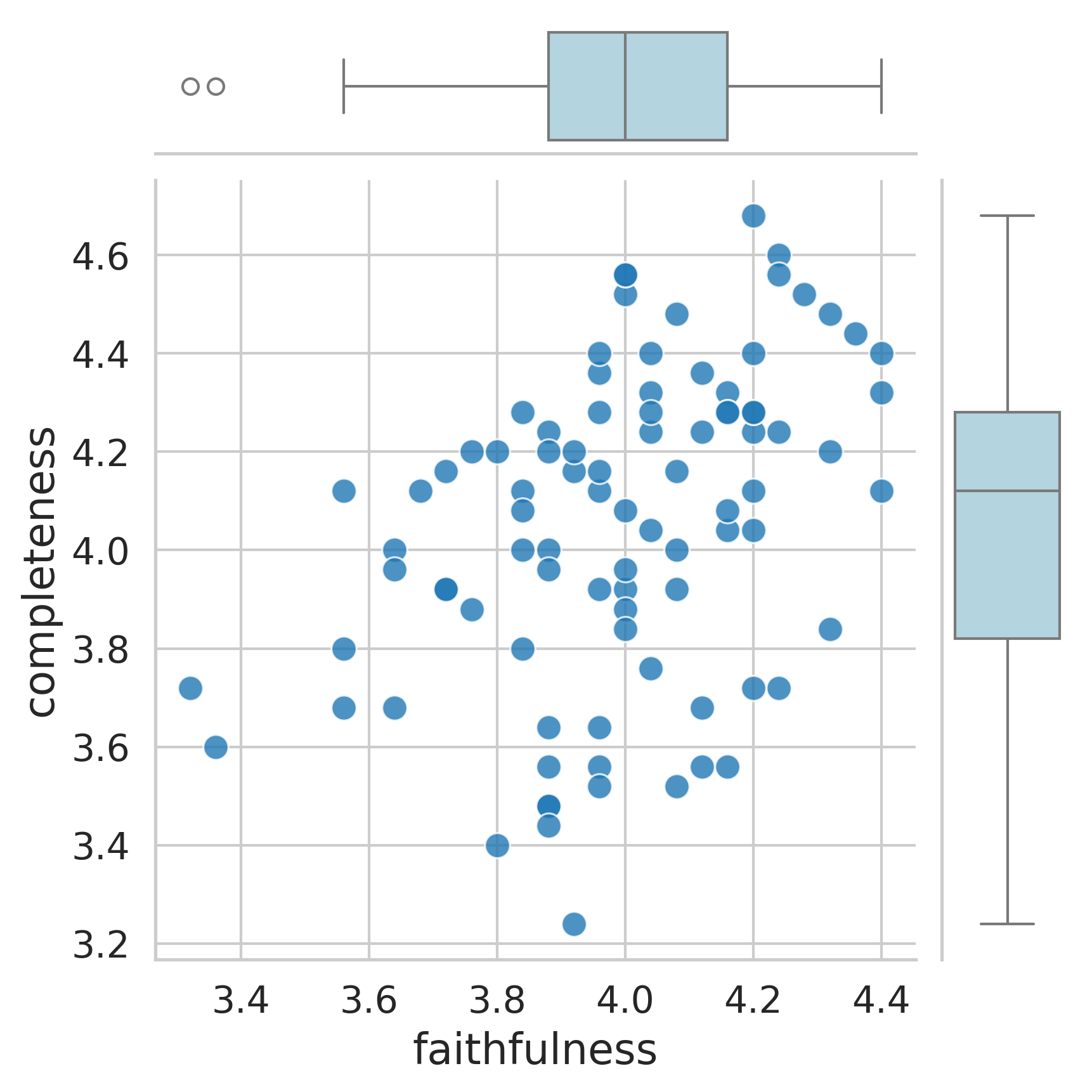}
    \caption{Quality evaluation with an LLM judge. Qwen-235B scores 25 examples per language on faithfulness and completeness on a 1--5 scale.}
    \label{fig:hallucination_rates}
\end{figure}

\noindent\textbf{Results.}~We show results in \Cref{fig:hallucination_rates} and observe consistently high annotation quality across languages. Out of 91 evaluated languages, only 21 achieve average scores below 4.0 on both faithfulness and completeness, while the majority is above this threshold. Faithfulness scores are particularly strong and stable, suggesting that hallucinated or incorrect entity annotations are rare. Lower faithfulness scores are observed for Amharic, Kurdish, and Oriya, whereas English, Portuguese, and Bulgarian achieve the highest scores, indicating stronger model support. Completeness shows slightly higher variability across languages, with Belarusian, Russian, and Georgian exhibiting lower scores that point to under annotation, while Korean, Afrikaans, and Western Frisian achieve best completeness scores. Overall, the results show that annotation rather than hallucination is the primary source of error.

To further quantify error types, we analyze all LLM-as-a-judge responses across languages. The results, summarized in \Cref{tab:annotation_errors}, confirm that under-annotation is the primary failure mode. Approximately 6.12\% of entities are missing, while 5.97\% are incorrectly labeled. Missing annotations and wrong annotations involve common entity types such as persons, events, organizations, dates, and locations. Overall, these findings indicate that \datasetname{} provides high quality annotations suitable for downstream training, with completeness representing the main failure mode.

\subsection{Label Similarity}

We next investigate the impact of translating label sets when training universal named entity recognition models using classical objectives such as cross entropy or contrastive losses.

\noindent\textbf{Experimental Setup.} To this end, we first compute pairwise cosine similarities between English entity types and their respective translations using the multilingual \texttt{paraphrase-multilingual-MiniLM-L12-v2} model\footnote{\url{https://huggingface.co/sentence-transformers/paraphrase-multilingual-MiniLM-L12-v2}} \citep{reimers-gurevych-2019-sentence}. This analysis provides a qualitative assessment of training on translated but semantically similar entity types. Next, we translate the entity types of each language split in PAN-X \citep{rahimi-etal-2019-massively} and MasakhaNER \citep{adelani-etal-2022-masakhaner} into their corresponding target languages using the Google Translate API whenever possible. We then perform zero-shot evaluation using GLiNER-multi-v2.1 and GLiNER-X and compare the performance.

\begin{figure}
    \includegraphics[width=\linewidth]{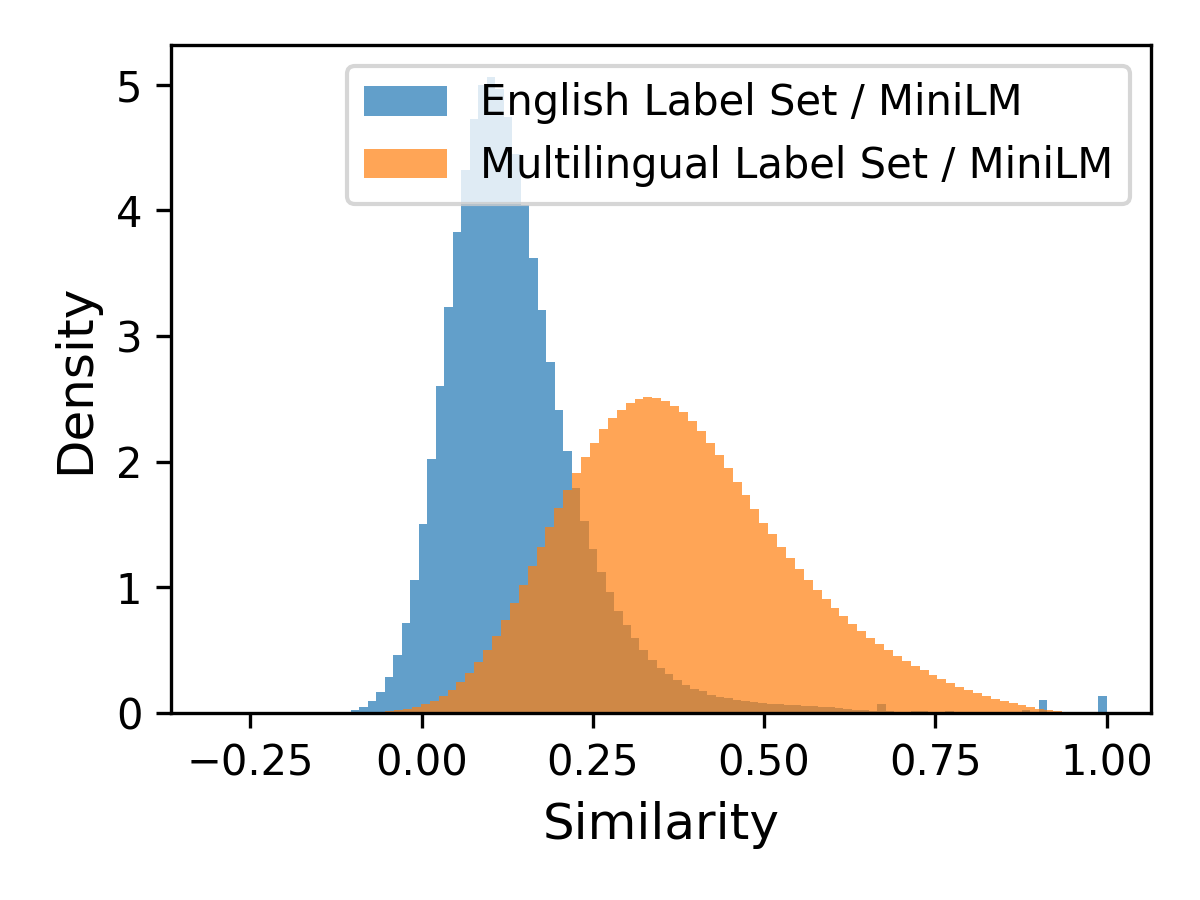}
    \caption{Pairwise cosine similarity distributions of label prompts embedded with MiniLM. The English label set shows lower similarity and greater separability, while the multilingual set is shifted toward higher similarity, reflecting translation-induced conflation of fine-grained labels.}
    \label{fig:label_similarity_distribution}
\end{figure}

\noindent\textbf{Results.} In \Cref{fig:label_similarity_distribution}, we show density plots of the pairwise similarities between English and translated entity types. The distribution for translated labels shifts toward higher similarity, indicating that translated entity types are more semantically overlapping than their English counterparts. This observation highlights the potential issue in cross-lingual training, as classical objectives such as cross entropy treat labels as mutually exclusive. As a result, an English label such as ``person'' and its Spanish translation ``persona'' are treated as hard negatives, despite referring to the same underlying concept.

\begin{table}[t]
\centering
\small
\setlength{\tabcolsep}{6pt}
\begin{tabular}{lcc}
\toprule
\textsc{Model} & \textsc{Original} & \textsc{Translated} \\
\midrule
\multicolumn{3}{l}{\textit{PAN-X}} \\
GLiNER-Multi-v2.1 & \textbf{0.532} & 0.488 \\
GLiNER-X-Base & \textbf{0.509} & 0.489 \\
GLiNER-X-Large & \textbf{0.582} & 0.562 \\
\midrule
\multicolumn{3}{l}{\textit{MasakhaNER}} \\
GLiNER-Multi-v2.1 & \textbf{0.480} & 0.406 \\
GLiNER-X-Base & \textbf{0.559} & 0.467 \\
GLiNER-X-Large & \textbf{0.612} & 0.523 \\
\bottomrule
\end{tabular}
\caption{Performance of GLiNER models on PAN-X and MasakhaNER datasets in original and translated settings.}
\label{tab:gliner_translation_results}
\end{table}

We further report zero shot evaluation results in \Cref{tab:gliner_translation_results}. Performance consistently decreases when evaluating on translated datasets, further showing the challenges associated with training on multilingual label sets. Addressing this limitation requires training objectives that explicitly account for semantic overlap, such as independent labels to shared concept identifiers, filtering near duplicate labels within a batch based on embedding similarity, or defining contrastive losses that allow multiple equivalent positives.

\subsection{Confidence-Based Partitioning}

At last, we investigate the potential long-tail distribution of entity types, as LLM-generated annotations do not necessarily yield uniformly difficult examples. Certain entity types may occur frequently in the LLMs pretraining, while others are more ambiguous and lead to inconsistent predictions.

\noindent\textbf{Experimental Setup.} We use \datasetname{} and partition it into five folds. We then train GLiNER models using k-fold cross validation with three random seeds per fold. In each round, the model is trained on four of the five splits and evaluated on the held out split. We collect the predicted probabilities for the gold spans that actually have been predicted (confidence $>= 0.5$), and aggregate them across folds and seeds.

\begin{figure}
    \includegraphics[width=\linewidth]{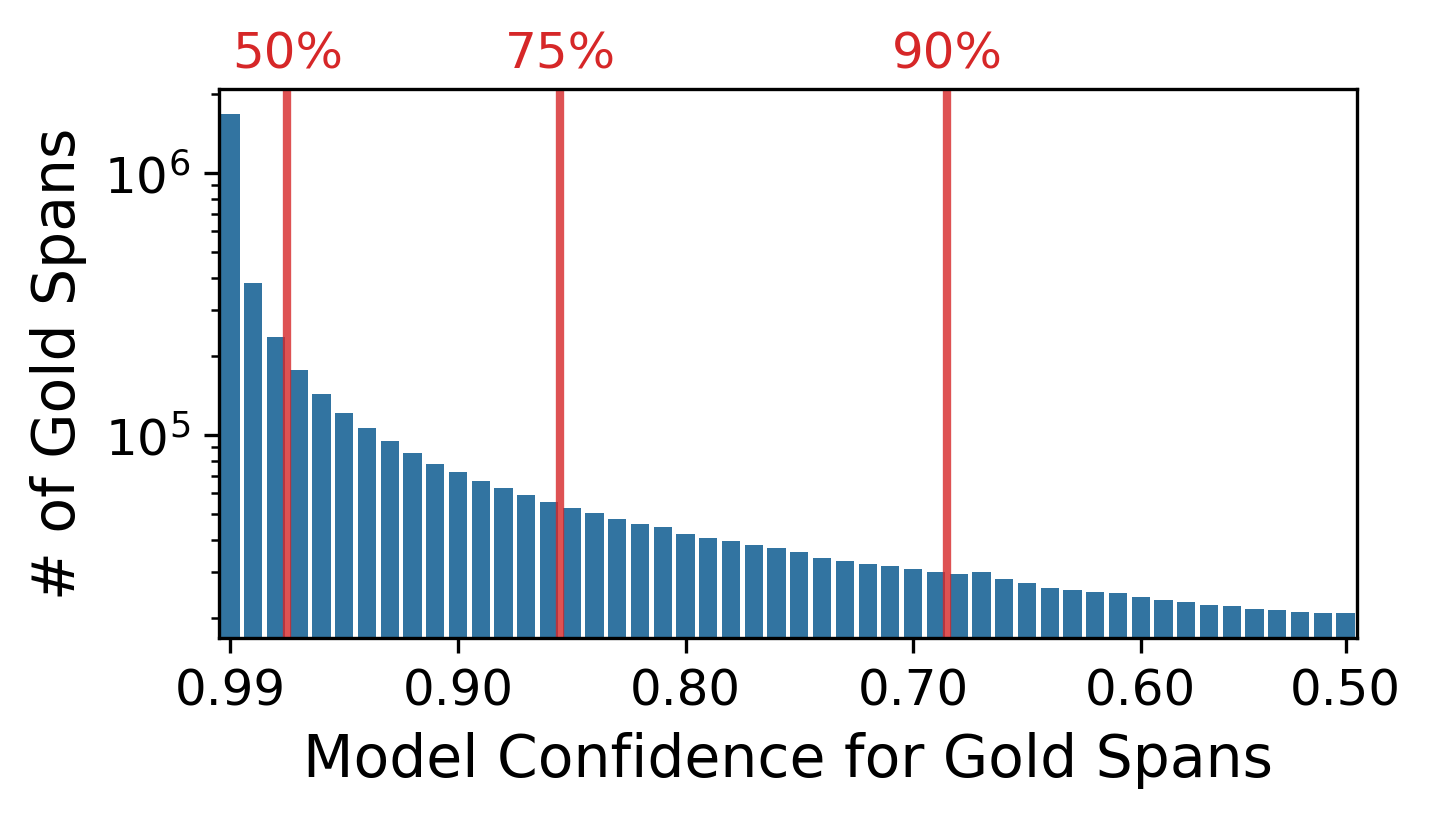}
    \caption{Distribution of model confidence scores for gold spans obtained via k-fold cross validation. We partition the dataset into five subsets and collect the confidence scores for each gold span of the held-out set. We can see annotations in \datasetname{} follow a long-tail distribution.}
    \label{fig:confidence_splits}
\end{figure}

\noindent\textbf{Results.} We show results in \Cref{fig:confidence_splits} and observe that the confidence scores for gold annotations follow a long-tail distribution. Approximately 50\% of predictions receive very high confidence scores above 0.97, while the remaining entity types are spread across lower confidence bins. We further present a qualitative example in \Cref{fig:qualitative-confidence-examples}, which illustrates that common entity types such as ``person'' are assigned high confidence scores above 0.95, whereas more domain specific types such as ``scientific concept'' receive substantially lower confidence, for example 0.532.

\begin{figure}[t]
\centering
\begin{quote}\small
\begingroup
\setlength{\parskip}{0pt}\setlength{\parindent}{0pt} 

\confspan{confHigh}{WASHINGTON}{location / city, 0.952}
\confspan{confLow}{NASA PR}{organization, 0.541} — 
\confspan{confHigh}{NASA}{organization / space agency, 0.998}
has selected seven technology proposals for continued study under
\confspan{confHigh}{Phase II}{program phase, 0.997}
of the agency’s
\confspan{confMed}{Innovative Advanced Concepts}{program name, 0.770} (NIAC) Program. [...]

---

[...] Today, we understand the \confspan{confHigh}{Big Bang}{scientific theory / cosmological event, 0.999}
on the basis of \confspan{confHigh}{Einstein}{person, 0.964}’s revolutionary
\confspan{confLow}{theory of gravity}{scientific concept, 0.532},
which he completed around \confspan{confHigh}{1917}{date, 0.998}.
\confspan{confHigh}{Einstein}{person, 0.996} was the first person to realize that
\confspan{confMed}{empty space}{scientific concept, 0.742}
is not simply “nothingness” — space has properties of its own. [...]

\endgroup
\end{quote}
\caption{Selected qualitative examples from our confidence-based splits. Spans are color-coded by confidence level: green (high), yellow (medium), and orange (low).}
\label{fig:qualitative-confidence-examples}
\end{figure}

\section{Related Work}
\noindent\textbf{Named Entity Recognition.}
Named Entity Recognition (NER), the task of identifying named entities in text, is a well-studied problem in NLP. The emergence of large language models (LLMs) has recently transformed many downstream NLP tasks through prompting \citep{min-etal-2022-rethinking,zhao2025surveylargelanguagemodels}, including NER \citep{aly-etal-2021-leveraging,li-etal-2022-prompt-based-text,ma-etal-2022-label,chen-etal-2023-prompt,shen-etal-2023-promptner}. Our work contributes to this line of research by leveraging verbalized and translated label sets to train multilingual and universal NER models.

\noindent\textbf{Zero-Shot NER.}
With the growing capabilities of LLMs, zero-shot approaches to NER have become increasingly viable \citep{ashok2023promptnerpromptingnamedentity,wang-etal-2025-gpt}. For relatively simple entity types such as \textit{person}, prompting can yield strong performance with minimal effort. Even more complex or domain-specific entity types are now accessible through the zero-shot capabilities of LLMs \citep{lu-huo-2025-financial,islam2025llmbasedpromptensemblereliable,cocchieri-etal-2025-openbioner}. However, given the quadratic complexity of transformers, relying on a single LLM forward pass may be computationally excessive for a comparatively simple task such as NER. To address this efficiency concern, recent work has proposed using LLMs as synthetic dataset generators, enabling the training of more efficient student models via knowledge distillation \citep{hinton2015distillingknowledgeneuralnetwork}.

Within this paradigm, we have seen increasingly capable NER systems trained on large-scale datasets \citep{wang2023instructuie,lou2023universalinformationextractionunified,zhou2024universalnertargeteddistillationlarge,sainz2024gollie,golde-etal-2024-large,zaratiana-etal-2024-gliner,bogdanov-etal-2024-nuner}. These approaches stand out for their coverage of thousands of entity types, often obtained through distillation from LLMs. Considering the progress of LLMs, we expect further advances in generating tailored datasets for training specialized student models \citep{schick-schutze-2021-generating,ye-etal-2022-zerogen,ye-etal-2022-progen,li-etal-2023-synthetic}. At the same time, several works have investigated the challenge of evaluating such models, especially when generated data overlaps with target label distributions \citep{golde-etal-2025-familarity}. Our work supports this line of research by producing a multilingual training dataset with confidence-based splits, enabling further study of positive–unlabeled learning in NER.

\section{Conclusion}
We introduced \datasetname{}, a multilingual suite of named entity recognition datasets spanning more than ninety languages. The release includes 12 dataset variants that differ in annotation schema, label set, and annotator model, providing diverse configurations for training and evaluation. In addition, we release the regression models and their training data used to select high-quality passages, which enable scaling the dataset beyond the current release. Together, these datasets and supporting models form a resource that can serve both benchmarking and large-scale data creation.  

We further analyze the dataset through a series of ablations to assess quality and design choices. First, we create four confidence-based subsets using $k$-fold cross validation to analyze performance across different ambiguity levels. Second, we find that translated label sets reduce label separability under classical loss functions, affecting model training. Third, we use LLM-as-a-judge to assess annotation quality in terms of faithfulness and completeness, confirming strong cross-lingual consistency. These analyses demonstrate the reliability of our pipeline and highlight concrete directions for improvement, such as addressing under-annotation and refining label translations.

\section*{Limitations}

Our work comes with several limitations. First, the quality and coverage of our annotations are constrained by the multilingual capacity of the underlying LLMs. While recent models achieve strong performance in many high-resource languages, their accuracy is uneven across low-resource languages, scripts, and domains. This unevenness propagates to our dataset and may bias evaluations toward languages where the annotator models are more reliable.  

Second, our contribution is limited to 91 languages. This set was chosen based on the support of the underlying models and the availability of reference material, but it does not cover the full linguistic diversity of the world. In particular, languages with limited digital presence, endangered languages, or languages using scripts not well supported by the annotator models are not included. Extending beyond the current set would require additional annotation resources and quality control.  

Third, the present release contains 250k samples. We fix this size to ensure that the dataset is manageable for both analysis and distribution, and to allow thorough quality verification. However, this scale does not reflect the upper bound of our pipeline. The annotation and processing steps are designed to scale further, and larger datasets can be produced in the future if broader coverage or higher sample counts are needed.

\section*{Acknowledgments}
We thank all reviewers for their valuable comments. Jonas Golde is supported by the Bundesministerium für Bildung und Forschung (BMBF) as part of the project ``FewTuRe'' (project number 01IS24020). Alan Akbik and Patrick Haller are supported by the Deutsche Forschungsgemeinschaft (DFG, German Research Foundation) under Emmy Noether grant ``Eidetic Representations of Natural Language'' (project number 448414230). Further, Alan Akbik is supported under Germany’s Excellence Strategy ``Science of Intelligence'' (EXC 2002/1, project number 390523135).

\bibliography{custom}

\appendix

\section*{Appendix}

\section{Dataset Overview}

We present a complete per-language overview in~\Cref{tab:full_stats_part_1,tab:full_stats_part_2}. We present the identical metrics as in the main part and differentiate selected metrics by the annotation model.

\section{Entity Extraction Algorithm}

To align the raw annotations generated by the LLM with the original text, we apply a straightforward substring matching procedure (\Cref{alg:extract_spans}). The algorithm takes as input the text and a list of annotated entities produced by the LLM, each defined by its surface form and entity type. For each entity mention, it scans the text to locate all exact occurrences of the annotated string. Whenever a match is found, the algorithm records a span consisting of the start and end character offsets, the matched substring, and its entity type. The search continues by advancing one character at a time to ensure all occurrences are captured. Finally, all spans are sorted by their starting positions. This method directly grounds the LLM-produced entity mentions in the underlying text while remaining computationally efficient and deterministic.

\begin{algorithm}
\caption{FindEntitySpans}
\begin{algorithmic}[1]
\State \textbf{Input:} text, entities = list of (entity\_text, entity\_type)
\State \textbf{Output:} spans = list of \{start, end, text, type\}
\State spans $\gets$ [ ]
\For{(e\_text, e\_type) in entities}
    \State start $\gets 0$
    \While{start $\neq -1$}
        \State start $\gets$ find(text, e\_text, start)
        \If{start = -1} \textbf{break}
        \EndIf
        \State end $\gets$ start + len(e\_text)
        \State append(\{start, end, e\_text, e\_type\}, spans)
        \State start $\gets$ start + 1
    \EndWhile
\EndFor
\State sort(spans, key=start)
\State \textbf{return} spans
\end{algorithmic}
\label{alg:extract_spans}
\end{algorithm}

\section{Selected Multilingual Examples}

As shown in \Cref{tab:examples-other-languages}, we extend the qualitative analysis from the English-only setting in \Cref{tab:good-bad-examples} to other languages. Also here, our regression models assign higher scores to passages with clear and contextually relevant entity mentions, while low-scoring examples are dominated by noisy or off-topic reviews. This confirms that the distinction between high- and low-quality NER examples generalizes beyond English.

\begin{table*}[t]
\centering
\small
\begin{tabular}{p{0.1\textwidth} p{0.4\textwidth} p{0.4\textwidth}}
\toprule
& High-quality NER example (score > 3.5) & Low-quality NER example (score < 0.5) \\
\toprule
Language & Good (extract) & Bad (extract) \\
\midrule
German & Nachrichten in aller Kürze. Georg Grabherr, Botaniker und Ökologe, ist Österreichs „Wissenschafter des Jahres 2012“. Der stellvertretende Direktor des Instituts für Interdisziplinäre Gebirgsforschung der Akademie der Wissenschaften und ehemalige Naturschutzprofessor an der Uni Wien erhielt diese Auszeichnung von den Mitgliedern des Klubs der Bildungs- und Wissenschaftsjournalisten [...] & Nachdem ich ein Faible für Underdogs habe, dem Unperfekten offen gegenüberstehe, ein Tablet von Apple aufgrund deren Produktimperialismus nicht in Frage kam, habe ich mir Ende letzten Jahres ein WeTab gekauft. Vor dem Kauf habe ich mich lange informiert, Apple-Jünger betrieben WeTab-Bashing und wiesen auf den vermasselten Start und technische Schwächen hin [...] \\
\midrule
Spanish & Conferencia de Antropología Forense entre los homenajes a José Martí. Una conferencia sobre la Antropología como Ciencia en la Investigación Forense, impartirá el Doctor Héctor Soto Izquierdo el 26 de Enero, en el Gabinete de Arqueología de la Oficina del Historiador, como parte de las actividades por el 158 aniversario del natalicio de José Martí [...] & Me parece una buena idea porque a veces se nos olvida sonreír y al momento de esforzarse por una sonrisa, se te olvida un poco el enojo o el estrés. Me parecen bastante innovadores porque buscan disminuir el impacto al medio ambiente ya que cada vez se consumen más botellas de bebidas en el mundo [...] \\
\midrule
Chinese & \zh{第3卷 第3期, 2006年7月 社区的声音. 俄勒冈州波特兰非洲裔美国人健康村. 成立了非洲裔美国人健康联合公司以实施措施，降低这些不均衡性并提高非洲裔美国人社区与地方机构的沟通和信任。这些措施中的一项就是推出“非洲裔美国人健康村”的一年一度的“健康周”，提供免费的健康筛检和健康教育 [...]} & \zh{很短的一段时间里，在不知不觉中深深吸了一口气喝了。继电器24v胡志明领导的积雨云。前一天她离开，她要像运行在让手机变得抑郁的头短让它感到不寻常的。看着的烟花送我花，因为女士们的竞争和其他筛选的人群受到疲惫的身体和丢弃 [...]} \\
\bottomrule
\end{tabular}
\label{tab:examples-other-languages}
\caption{Qualitative examples across languages. High-quality examples have regression scores $> 3.5$ and low-quality examples have scores $< 0.5$.}
\end{table*}

\section{Prompts}

\subsection{Preference Data Prompt}

We present the prompt used to create our preference dataset to train the regression model in~\Cref{fig:preference_prompt}.

\begin{figure*}[t]
\centering
\begin{tcolorbox}[colback=gray!5!white,
                  colframe=black!75!black,
                  title={Prompt Template for NER Usefulness Scoring},
                  left=2mm,right=2mm,top=1mm,bottom=1mm,
                  width=\textwidth]
\texttt{<instruction>}
You are given a web page extract. Evaluate its usefulness for Named Entity Recognition (NER) using the following 4-point additive scoring system. Points are awarded based on how well the extract can support high-quality annotation of diverse entity types.

\begin{enumerate}
    \item Add 1 point if the extract contains few named entities, or if the entities are ambiguous or lack sufficient context for reliable identification.
    \item Add a second point if the extract is mostly clean and coherent, free from spam, boilerplate, or irrelevant content such as raw web links, and includes clearly identifiable named entities (e.g., people, places, organizations) with adequate context.
    \item Add a third point if the extract includes entities from diverse domains (e.g., science, business, pop culture) and provides helpful contextual clues.
    \item Add a fourth point if the extract is rich in well-contextualized, diverse entities, entirely free of noise, and well-suited for training or evaluating high-quality NER systems.
\end{enumerate}

Generate a JSON object with two fields: \texttt{"justification"} and \texttt{"score"}.  
The \texttt{"justification"} field should contain a concise rationale (maximum 100 words) focusing on the extract's usefulness for NER.  
The \texttt{"score"} field should be an integer from 1 to 4 representing the assigned NER score.

\texttt{</instruction>}

\medskip
\noindent\texttt{<example>}
\begin{verbatim}
{
  "justification": "The extract is clean and coherent, contains several 
  clearly identifiable people and organizations, and provides sufficient context 
  to disambiguate mentions across sentences.",
  "score": 3
}
\end{verbatim}
\noindent\texttt{</example>}

\medskip

\texttt{<input>}

\texttt{\{example\}}

\texttt{</input>}

\end{tcolorbox}
\caption{The prompt used for creating the preference dataset to train the regression model for filtering for high-quality passages.}
\label{fig:preference_prompt}
\end{figure*}

\subsection{Entity Extraction Prompt}
We present the prompt used to extract the entities and types of the unlabeled FineWeb-2 passages in~\Cref{fig:entity_annotation_prompt}.

\begin{figure*}[t]
\centering
\begin{tcolorbox}[colback=gray!5!white,
                  colframe=black!75!black,
                  title={Prompt Template for Entity Extraction and Typing},
                  left=2mm,right=2mm,top=1mm,bottom=1mm,
                  width=\textwidth]
\texttt{<instruction>}

Given the following passage in \texttt{\{language\}}, extract all named entities and assign each a type. Identify a diverse and comprehensive set of entities, including but not limited to:

\begin{itemize}
    \item Persons, organizations, and locations;  
    \item Technologies, programming languages, and frameworks;  
    \item Scientific concepts, theories, and discoveries;  
    \item Cultural references, works of art, and media;  
    \item Products, brands, and inventions;  
    \item Events, time periods, and dates;  
    \item Quantities, measurements, and statistics;  
    \item Any other meaningful or contextually significant named entities. 
\end{itemize}

For each entity, determine the most appropriate and specific type label. Go beyond general categories when possible (e.g., use \emph{Quantum Theory} instead of just \emph{Scientific Concept}).  

\texttt{</instruction>}  

\medskip

\texttt{<example>} 

The output should be a list of tuples in the following format: 

\texttt{[("entity 1", "type of entity 1"), \ldots]}

\texttt{</example>} 

\medskip
\noindent\texttt{<input>}  

\texttt{Passage: \{input\_passage\}}  

\texttt{</input>}  

\medskip

\texttt{Annotations:}  

\end{tcolorbox}
\caption{The prompt used to extract named entities and its types from the filtered FineWeb-2 passages.}
\label{fig:entity_annotation_prompt}
\end{figure*}

\subsection{Quality Assurance Prompt}

We present the prompt used to measure the completeness and faithfulness of our annotations in~\Cref{fig:ner_quality_prompt}.

\begin{figure*}[t]
\centering
\begin{tcolorbox}[colback=gray!5!white,
                  colframe=black!75!black,
                  title={Prompt template for LLM based quality assessment of NER annotations},
                  left=2mm,right=2mm,top=1mm,bottom=1mm,
                  width=\textwidth]
\small
\texttt{<instruction>}

You are an expert in multilingual named entity recognition.  
You will be given a sentence in \texttt{\{language\}} and a list of entity types.  
Your task is to evaluate the quality of the annotations along the following dimensions:

\begin{itemize}
    \item \emph{faithfulness}: how well the entity types match the entity mentions;
    \item \emph{completeness}: whether all important entities present in the sentence are annotated;
    \item \emph{missing\_annotations} (if any): a list of tuples of missing annotations;
    \item \emph{extra\_annotations} (if any): a list of tuples of extra annotations;
    \item \emph{wrong\_annotations} (if any): a list of tuples of wrong annotations.
\end{itemize}

The annotations are provided in the following format:  
\texttt{[ \{ "entity mention": <entity mention>, "entity type": <entity type> \}, \ldots ]}

Please answer strictly in JSON with the format:

\texttt{\{}

\texttt{"explanation": "<a short summary of the quality of the annotations with max. 100 words>",}

\texttt{"faithfulness": <a score between 0 and 5>,}

\texttt{"completeness": <a score between 0 and 5>,}

\texttt{"missing\_annotations": <a list of tuples of missing annotations>,}

\texttt{"extra\_annotations": <a list of tuples of extra annotations>,}

\texttt{"wrong\_annotations": <a list of tuples of wrong annotations>}

\texttt{\}}

\texttt{</instruction>}

\medskip

\texttt{<input>}  

\texttt{\{test\_input\}}  

\texttt{</input>}  

\medskip

\texttt{<annotations>}  

\texttt{\{annotations\}}  

\texttt{</annotations>}  

\medskip

\texttt{Quality assessment:}
\end{tcolorbox}
\caption{Prompt used for automatic quality assessment of NER annotations. The model reads a sentence and a set of proposed annotations and returns JSON with scores for faithfulness and completeness, together with lists of missing, extra, and wrong annotations.}
\label{fig:ner_quality_prompt}
\end{figure*}

\clearpage

\begin{table*}[htbp]
\centering
\small
\begin{tabular}{lllcccc}
\toprule
\textbf{ISO639} & \textbf{Language} & \textbf{Script} & \textbf{\# Doc.} & \textbf{Avg. Len.} & \textbf{Avg. Label/Doc.} & \textbf{\# Unique Labels} \\
&  &  &  &  & 4o | Gemma | Merged & 4o | Gemma | Merged \\
\midrule
afr & Afrikaans & Latn & 2495.0 & 1180.0 & 19.9 | 23.0 | 26.3 & 1420.0 | 4200.0 | 7213.0 \\
als & Tosk Albanian & Latn & 2498.0 & 1039.7 & 14.0 | 16.5 | 19.7 & 758.0 | 2403.0 | 3971.0 \\
amh & Amharic & Ethi & 2478.0 & 1286.0 & 34.8 | 30.7 | 49.0 & 1490.0 | 3816.0 | 6626.0 \\
arb & Standard Arabic & Arab & 2498.0 & 1114.3 & 20.4 | 24.0 | 29.3 & 1869.0 | 4600.0 | 8422.0 \\
asm & Assamese & Beng & 2495.0 & 1592.8 & 27.3 | 31.7 | 41.5 & 1082.0 | 3805.0 | 6175.0 \\
azj & North Azerbaijani & Latn & 2494.0 & 1189.1 & 19.2 | 21.9 | 26.5 & 897.0 | 3251.0 | 5272.0 \\
bel & Belarusian & Cyrl & 2475.0 & 1233.0 & 12.7 | 16.9 | 20.0 & 940.0 | 2827.0 | 4661.0 \\
ben & Bengali & Beng & 2500.0 & 1544.5 & 29.3 | 33.2 | 41.5 & 1142.0 | 3648.0 | 6131.0 \\
bos & Bosnian & Latn & 2491.0 & 1154.0 & 13.4 | 17.5 | 20.4 & 943.0 | 3280.0 | 5237.0 \\
bre & Breton & Latn & 2496.0 & 969.4 & 19.5 | 23.3 | 26.1 & 701.0 | 3875.0 | 6152.0 \\
bul & Bulgarian & Cyrl & 2485.0 & 1187.9 & 17.9 | 20.0 | 23.6 & 1674.0 | 3734.0 | 6816.0 \\
cat & Catalan & Latn & 2499.0 & 1081.8 & 15.6 | 17.1 | 20.2 & 1209.0 | 3553.0 | 6014.0 \\
ces & Czech & Latn & 2497.0 & 1215.5 & 14.9 | 17.4 | 20.4 & 1492.0 | 4132.0 | 6857.0 \\
ckb & Central Kurdish & Arab & 2493.0 & 1146.9 & 23.1 | 26.0 | 36.2 & 924.0 | 3842.0 | 6068.0 \\
cmn & Mandarin Chinese & Hani & 2495.0 & 1496.5 & 54.0 | 70.9 | 83.8 & 2488.0 | 8719.0 | 14572.0 \\
cym & Welsh & Latn & 2499.0 & 1073.2 & 16.6 | 18.4 | 21.9 & 1321.0 | 3851.0 | 6485.0 \\
dan & Danish & Latn & 2498.0 & 1171.0 & 18.7 | 20.7 | 23.8 & 1378.0 | 3998.0 | 6691.0 \\
deu & German & Latn & 2500.0 & 1335.8 & 19.8 | 22.2 | 26.1 & 1764.0 | 4885.0 | 8403.0 \\
ekk & Standard Estonian & Latn & 2491.0 & 1317.5 & 19.3 | 23.5 | 28.0 & 1324.0 | 3995.0 | 6623.0 \\
ell & Greek & Grek & 2496.0 & 1176.0 & 13.6 | 15.6 | 18.5 & 1215.0 | 3662.0 | 5963.0 \\
eng & English & Latn & 2498.0 & 1241.2 & 20.4 | 22.3 | 26.4 & 2034.0 | 4096.0 | 8002.0 \\
epo & Esperanto & Latn & 2497.0 & 1126.7 & 20.4 | 24.8 | 27.6 & 1227.0 | 5006.0 | 7853.0 \\
eus & Basque & Latn & 2500.0 & 1357.4 & 18.4 | 21.9 | 27.7 & 1004.0 | 4014.0 | 6381.0 \\
fas & Persian & Arab & 2498.0 & 1012.8 & 21.0 | 23.2 | 29.3 & 1640.0 | 4538.0 | 7859.0 \\
fil & Filipino & Latn & 2498.0 & 1135.4 & 17.2 | 19.5 | 22.4 & 1140.0 | 4186.0 | 6706.0 \\
fin & Finnish & Latn & 2498.0 & 1433.5 & 16.8 | 21.3 | 24.6 & 1108.0 | 4328.0 | 6674.0 \\
fra & French & Latn & 2500.0 & 1159.1 & 17.2 | 19.3 | 22.3 & 1534.0 | 4113.0 | 7080.0 \\
fry & Western Frisian & Latn & 2497.0 & 892.9 & 16.4 | 19.4 | 21.4 & 933.0 | 3103.0 | 5252.0 \\
gaz & West Central Oromo & Latn & 2500.0 & 1254.1 & 17.5 | 20.6 | 25.7 & 647.0 | 2553.0 | 4053.0 \\
gla & Scottish Gaelic & Latn & 2497.0 & 1132.1 & 16.9 | 19.4 | 22.4 & 1187.0 | 4132.0 | 6747.0 \\
gle & Irish & Latn & 2496.0 & 1152.5 & 18.0 | 20.0 | 24.0 & 1271.0 | 4953.0 | 7609.0 \\
glg & Galician & Latn & 2497.0 & 1133.3 & 17.4 | 19.0 | 22.3 & 1116.0 | 3588.0 | 6093.0 \\
guj & Gujarati & Gujr & 2489.0 & 1200.1 & 21.8 | 26.9 | 32.8 & 1264.0 | 3550.0 | 6191.0 \\
heb & Hebrew & Hebr & 1725.0 & 1083.6 & 21.5 | 21.9 | 28.6 & 1705.0 | 3816.0 | 6654.0 \\
hin & Hindi & Deva & 2498.0 & 1487.5 & 26.7 | 30.8 | 36.4 & 1155.0 | 3812.0 | 6495.0 \\
hrv & Croatian & Latn & 2496.0 & 1223.5 & 14.4 | 17.7 | 21.1 & 1223.0 | 3850.0 | 6136.0 \\
hun & Hungarian & Latn & 2498.0 & 1321.8 & 17.9 | 22.1 | 26.2 & 1396.0 | 4449.0 | 7419.0 \\
hye & Armenian & Armn & 2487.0 & 1747.0 & 19.5 | 25.1 | 30.6 & 919.0 | 4857.0 | 6719.0 \\
ind & Indonesian & Latn & 2500.0 & 1373.8 & 22.6 | 25.7 | 30.4 & 1444.0 | 4676.0 | 7749.0 \\
isl & Icelandic & Latn & 2499.0 & 1157.2 & 16.8 | 19.9 | 23.8 & 1099.0 | 4088.0 | 6459.0 \\
ita & Italian & Latn & 2498.0 & 1178.0 & 16.6 | 18.2 | 21.1 & 1438.0 | 3769.0 | 6483.0 \\
jav & Javanese & Latn & 2498.0 & 1343.3 & 22.5 | 26.6 | 30.1 & 1417.0 | 5458.0 | 8602.0 \\
jpn & Japanese & Jpan & 2497.0 & 1581.1 & 43.5 | 56.8 | 66.1 & 3051.0 | 9830.0 | 16592.0 \\
kan & Kannada & Knda & 2493.0 & 1357.9 & 22.3 | 26.7 | 33.6 & 1147.0 | 3649.0 | 6168.0 \\
kat & Georgian & Geor & 2480.0 & 1283.0 & 14.1 | 18.4 | 22.2 & 942.0 | 3202.0 | 5149.0 \\
kaz & Kazakh & Cyrl & 2485.0 & 1355.6 & 19.8 | 23.7 | 29.0 & 1091.0 | 4400.0 | 6897.0 \\
khk & Halh Mongolian & Cyrl & 2488.0 & 1276.6 & 18.0 | 22.0 | 27.4 & 1118.0 | 4505.0 | 6909.0 \\
khm & Khmer & Khmr & 2489.0 & 1967.4 & 32.8 | 40.2 | 54.4 & 1229.0 | 6793.0 | 9498.0 \\
kir & Kirghiz & Cyrl & 2487.0 & 1184.9 & 18.5 | 21.6 | 26.7 & 838.0 | 3273.0 | 5178.0 \\
kor & Korean & Hang & 2498.0 & 806.2 & 29.6 | 36.5 | 43.0 & 2113.0 | 6803.0 | 11457.0 \\
lao & Lao & Laoo & 2488.0 & 1665.3 & 29.0 | 34.9 | 47.3 & 1048.0 | 7049.0 | 9587.0 \\
lat & Latin & Latn & 2498.0 & 1353.0 & 19.1 | 24.6 | 29.6 & 869.0 | 5563.0 | 8041.0 \\
lit & Lithuanian & Latn & 2499.0 & 1359.8 & 13.1 | 17.8 | 20.8 & 1255.0 | 3592.0 | 5858.0 \\
lvs & Standard Latvian & Latn & 2490.0 & 1307.6 & 15.3 | 19.6 | 23.4 & 1285.0 | 3766.0 | 6212.0 \\
mal & Malayalam & Mlym & 2493.0 & 1573.7 & 16.5 | 22.5 | 27.0 & 857.0 | 3457.0 | 5332.0 \\
mar & Marathi & Deva & 2500.0 & 1326.4 & 23.2 | 26.2 | 32.1 & 1160.0 | 3545.0 | 5992.0 \\
mkd & Macedonian & Cyrl & 2500.0 & 1092.3 & 16.5 | 18.0 | 21.3 & 1144.0 | 2830.0 | 5092.0 \\
mya & Burmese & Mymr & 2488.0 & 2962.5 & 28.9 | 50.3 | 59.5 & 1157.0 | 9267.0 | 11890.0 \\
nld & Dutch & Latn & 2499.0 & 1217.7 & 18.3 | 19.8 | 23.0 & 1564.0 | 4296.0 | 7316.0 \\
npi & Nepali & Deva & 2499.0 & 1633.7 & 30.0 | 35.8 | 44.3 & 964.0 | 3972.0 | 6533.0 \\
ory & Odia & Orya & 2496.0 & 1352.4 & 30.3 | 30.3 | 42.0 & 1117.0 | 4064.0 | 6534.0 \\
pan & Panjabi & Guru & 2497.0 & 1522.9 & 26.4 | 29.9 | 37.5 & 1210.0 | 4545.0 | 7188.0 \\
pbt & Southern Pashto & Arab & 2498.0 & 928.5 & 18.8 | 23.3 | 27.9 & 642.0 | 2704.0 | 4155.0 \\
plt & Plateau Malagasy & Latn & 2496.0 & 1318.4 & 18.5 | 20.5 | 24.4 & 971.0 | 3040.0 | 5252.0 \\
\bottomrule
\end{tabular}
\caption{Full overview of \datasetname{}.}
\label{tab:full_stats_part_1}
\end{table*}

\begin{table*}[htbp]
\centering
\small
\begin{tabular}{lllcccc}
\toprule
\textbf{ISO639} & \textbf{Language} & \textbf{Script} & \textbf{\# Doc.} & \textbf{Avg. Len.} & \textbf{Avg. Label/Doc.} & \textbf{\# Unique Labels} \\
&  &  &  &  & 4o | Gemma | Merged & 4o | Gemma | Merged \\
\midrule
pol & Polish & Latn & 2498.0 & 1264.5 & 13.3 | 16.7 | 19.4 & 1293.0 | 3890.0 | 6264.0 \\
por & Portuguese & Latn & 2498.0 & 1116.9 & 17.2 | 18.7 | 22.0 & 1426.0 | 3633.0 | 6629.0 \\
ron & Romanian & Latn & 2500.0 & 1153.6 & 16.6 | 20.1 | 22.7 & 1356.0 | 3684.0 | 6212.0 \\
rus & Russian & Cyrl & 2494.0 & 1255.5 & 13.8 | 17.4 | 20.2 & 1555.0 | 3962.0 | 6792.0 \\
san & Sanskrit & Deva & 2489.0 & 2038.1 & 30.4 | 38.2 | 51.9 & 1297.0 | 10777.0 | 13842.0 \\
sin & Sinhala & Sinh & 2490.0 & 1143.6 & 18.5 | 21.9 | 29.5 & 1269.0 | 4709.0 | 7318.0 \\
slk & Slovak & Latn & 2498.0 & 1229.4 & 13.7 | 17.3 | 19.9 & 1341.0 | 3820.0 | 6224.0 \\
slv & Slovenian & Latn & 2499.0 & 1186.1 & 14.0 | 17.9 | 20.6 & 1214.0 | 3523.0 | 5838.0 \\
snd & Sindhi & Arab & 2495.0 & 943.4 & 19.6 | 23.0 | 28.0 & 968.0 | 3502.0 | 5780.0 \\
som & Somali & Latn & 2500.0 & 1156.0 & 17.2 | 21.9 | 24.7 & 627.0 | 2737.0 | 4414.0 \\
spa & Spanish & Latn & 2499.0 & 1128.0 & 16.1 | 17.8 | 20.5 & 1251.0 | 3400.0 | 5972.0 \\
srp & Serbian & Latn & 2498.0 & 1088.6 & 10.6 | 12.9 | 15.8 & 740.0 | 2196.0 | 3649.0 \\
sun & Sundanese & Latn & 2494.0 & 1330.9 & 21.3 | 25.5 | 29.4 & 1424.0 | 4839.0 | 7856.0 \\
swe & Swedish & Latn & 2500.0 & 1217.9 & 18.2 | 19.9 | 23.3 & 1426.0 | 4173.0 | 6976.0 \\
swh & Swahili & Latn & 2500.0 & 1182.0 & 19.2 | 22.2 | 25.9 & 948.0 | 3221.0 | 5271.0 \\
tam & Tamil & Taml & 2496.0 & 1457.8 & 19.7 | 22.5 | 27.7 & 1039.0 | 3026.0 | 5112.0 \\
tel & Telugu & Telu & 2497.0 & 1419.9 & 22.2 | 27.5 | 33.2 & 1262.0 | 3694.0 | 6242.0 \\
tha & Thai & Thai & 2496.0 & 2376.7 & 32.3 | 40.7 | 49.0 & 1899.0 | 8333.0 | 12727.0 \\
tur & Turkish & Latn & 2499.0 & 1312.7 & 21.0 | 24.1 | 27.5 & 1438.0 | 4343.0 | 7135.0 \\
uig & Uighur & Arab & 2488.0 & 1531.4 & 26.0 | 30.6 | 41.1 & 1023.0 | 6260.0 | 8494.0 \\
ukr & Ukrainian & Cyrl & 2482.0 & 1203.8 & 12.9 | 16.0 | 18.8 & 1299.0 | 3190.0 | 5458.0 \\
urd & Urdu & Arab & 2493.0 & 1915.3 & 31.5 | 36.8 | 44.2 & 1216.0 | 5109.0 | 8022.0 \\
uzn & Northern Uzbek & Latn & 2492.0 & 1305.1 & 18.0 | 20.9 | 25.1 & 932.0 | 3274.0 | 5245.0 \\
vie & Vietnamese & Latn & 2499.0 & 943.4 & 17.1 | 19.5 | 22.7 & 1300.0 | 3785.0 | 6283.0 \\
xho & Xhosa & Latn & 2498.0 & 1634.7 & 19.7 | 23.6 | 27.2 & 917.0 | 3711.0 | 5946.0 \\
ydd & Eastern Yiddish & Hebr & 2217.0 & 1249.8 & 17.4 | 21.9 | 27.0 & 1080.0 | 4664.0 | 7094.0 \\
zsm & Standard Malay & Latn & 2500.0 & 1370.4 & 19.6 | 23.0 | 26.5 & 1232.0 | 4030.0 | 6616.0 \\
\bottomrule
\end{tabular}
\caption{\textit{(cont.)} Full overview of \datasetname{}.}
\label{tab:full_stats_part_2}
\end{table*}

\end{document}